\documentclass{article}
\PassOptionsToPackage{numbers, compress}{natbib}
\usepackage[final]{neurips_2024}
\usepackage{hyperref}   
\usepackage[title]{appendix}
\usepackage{mathtools}
\usepackage{booktabs}       
\usepackage{amsfonts}       
\usepackage{nicefrac} 
\usepackage{placeins}
\usepackage{thm-restate}
\usepackage{microtype}      
\usepackage{xcolor}
\usepackage{graphicx}
\usepackage{enumitem}
\usepackage{thmtools}
\usepackage{amsmath,amssymb,amsthm,bm}
\usepackage{subcaption,graphicx}
\usepackage{caption}
\usepackage{cleveref}
\usepackage{wrapfig}
\usepackage{stmaryrd}
\usepackage{multirow}
\usepackage{adjustbox}

\usepackage{thm-restate}

\theoremstyle{definition}

\theoremstyle{remark}

\renewcommand{\[}{\begin{eqnarray}}
\renewcommand{\]}{\end{eqnarray}}

\usepackage[ruled,vlined,linesnumbered]{algorithm2e}

\usepackage{etoolbox}


\SetKwInput{KwInput}{Input}                
\SetKwInput{KwOutput}{Output}
\usepackage{standalone}
\usepackage{tikz}
\usetikzlibrary{positioning}
\usetikzlibrary{calc,arrows,positioning, fit}

\usepackage{float}
\renewcommand{\[}{\begin{eqnarray}}
\renewcommand{\]}{\end{eqnarray}}

\usepackage{mathbbol}

\usepackage{thm-restate}

\newcommand{\R}{\mathbb{R}}

\title{Enhancing JEPAs with Spatial Conditioning: Robust and Efficient Representation Learning}
\author{Authors}

\author{
		Etai Littwin$^{1}$,\, Vimal Thilak$^{1}$, Anand Gopalakrishnan$^{2}$\thanks{Work done while interning at Apple.} \\
		$^1$ \textit{Apple} \\
	    $^2$ \textit{The Swiss AI Lab, IDSIA, USI \& SUPSI} \\
	}
 
\begin{document}

\maketitle

\begin{abstract}
Image-based Joint-Embedding Predictive Architecture (IJEPA) offers an attractive alternative to Masked Autoencoder (MAE) for representation learning using the Masked Image Modeling framework.
IJEPA drives representations to capture useful semantic information by predicting in latent rather than input space. 
However, IJEPA relies on carefully designed context and target windows to avoid representational collapse.
The encoder modules in IJEPA cannot adaptively modulate the type of predicted and/or target features based on the feasibility of the masked prediction task as they are not given sufficient information of both context and targets. 
Based on the intuition that in natural images, information has a strong spatial bias with spatially local regions being highly predictive of one another compared to distant ones. 
We condition the target encoder and context encoder modules in IJEPA with positions of context and target windows respectively. 
Our ``conditional'' encoders show performance gains on several image classification benchmark datasets, improved robustness to context window size and sample-efficiency during pretraining.
\end{abstract}

\section{Introduction}
 Masked Image Modeling (MIM) offers a scalable framework to learn representations from unlabelled data in a self-supervised manner by learning to predict masked regions given unmasked ones as context \citep{pathak2016contextef, xie2021simmim, bao2022beit, he2022masked, zhou2022image, baevski2022data2vec, baevski2022efficientssl, ijepa}.
A distinction can be drawn for models under this framework based on whether the targets are predicted in input space (pixels, words, sounds etc.) by MAEs \citep{he2022masked} or in latent space by JEPAs \cite{ijepa, bardes2024revisiting}. 
Recently, \citet{littwin2024jepa} suggest that JEPAs have an implicit bias for learning ``high-influence'' features compared to Masked Autoencoders (MAEs) which could explain their empirical success compared to MAEs.
However, JEPAs require careful selection of context and target windows (window size and distance of separation) to drive the representations to capture useful information (semantics) from input images for a variety of high-level downstream tasks like image classification as well as fine-grained tasks like object counting and depth prediction. 
Sub-optimal choice of context and target windows, i.e. pairs with low mutual information, potentially leads to representational collapse. 
Our work attempts to alleviate these limitations in JEPAs \citep{ijepa, bardes2024revisiting} --- improve representational quality to solve downstream tasks and robustness to masking hyperparameters for pretraining. 
\looseness=-1

\begin{figure*}[t]
	\centering
	\includegraphics[width=0.67\linewidth]{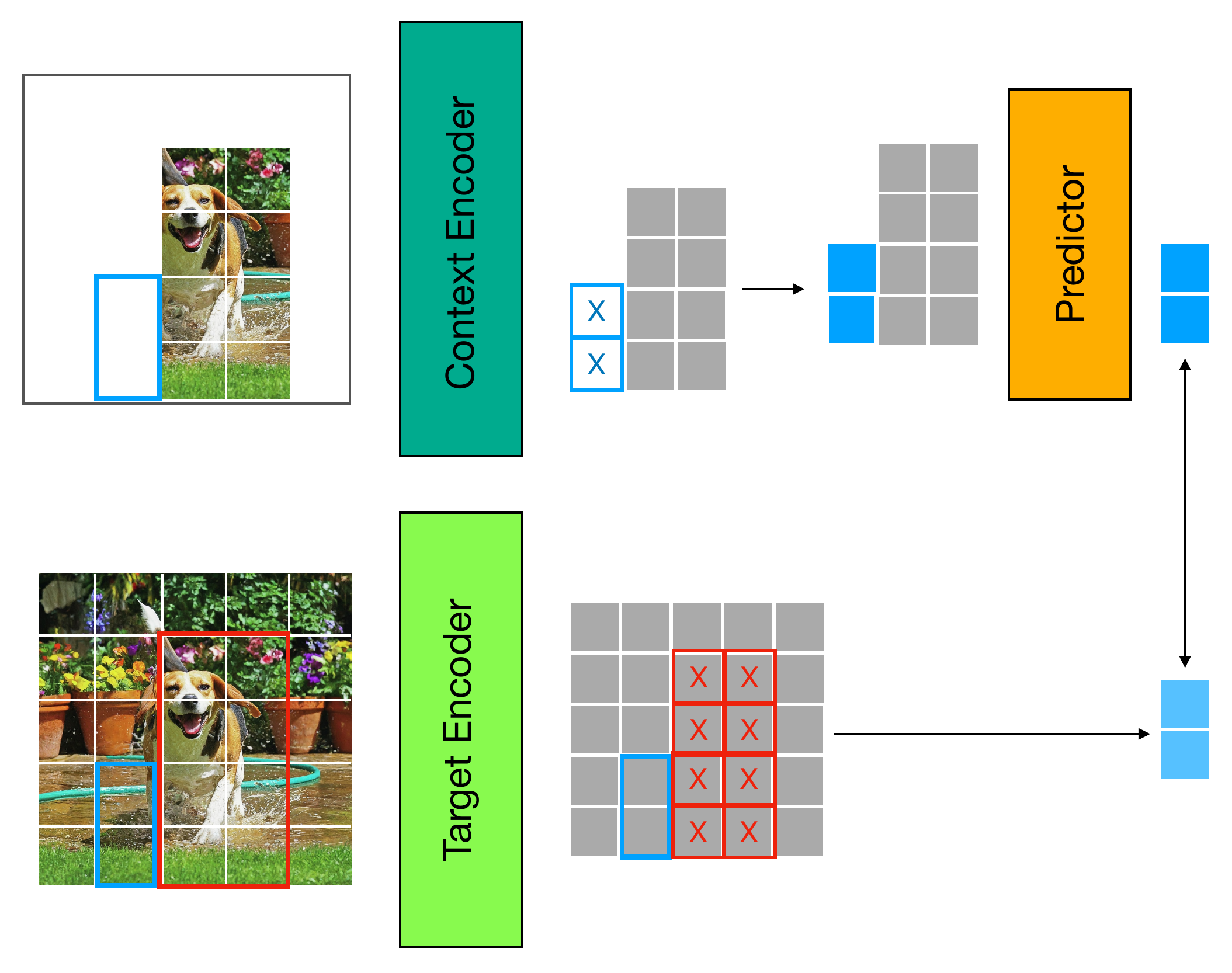}
	\caption{Conditioning the Context and Target Encoders in IJEPA with positions of the target (blue box) and context windows (red box) respectively. Patches marked with X indicate positional information while those with solid color fill indicate feature information is extracted at those locations.}
	\label{fig: model diagram}
\end{figure*}

In natural images, it is intuitive to expect nearby regions to be highly predictive of one another (high mutual information) compared to distant ones. 
The feasibility of the masked prediction task in JEPAs is linked to the mutual information between context and target windows. 
Consider the scene in \Cref{fig: model diagram} of the dog in the backyard, patches of grass co-occur with patches of flower pots but its plausible in other scenes for grass patches to co-occur with patches of sky, trees, water etc.
Therefore grass alone is not a highly predictive contextual feature for flower pots. 
On the other hand, patches from the same object (eg. dog), are highly predictive of each other as they co-occur almost always. 
Good choices for context and target masks in MIM require a careful balance of the amount of mutual information between image regions in the context and target windows. 
When the mutual information between image regions in the context and target windows is too low the prediction task is very challenging. 
This forces the encoders to extract only the most feasible set of features to predict from the target given a context leading to representational collapse in the limiting case. 
While if the mutual information is too high it becomes rather trivial resulting in the representations not capturing sufficiently abstract information from the input image. 

In JEPAs (eg. IJEPA \citep{ijepa}), the context and target encoders are given insufficient information about the prediction task as they do not have access to both context and target windows.
Therefore, the target encoder module cannot adaptively modulate target features (feedback signal) based on the feasibility of prediction to the context encoder. 
Without providing the context encoder and target encoder modules sufficient information of the masked prediction task, they can only extract highly predictable features from the context and target windows which could lead to representational collapse. 
Since predictability of information in natural images has a strong spatial bias as outlined above, providing information of sizes of context and target windows and the distance of separation could alleviate this issue. 
\looseness=-1

We incorporate this intuition in IJEPA \citep{ijepa} by conditioning the context encoder with positions of the target window and conversely the target encoder with positions of the context window.  
Given this additional information of spatial locations of target patches allows the context encoder to modulate the type of features to capture (low-level features $\rightarrow$ color, texture, shape or higher-level features $\rightarrow$ object categories) from the input image. 
Conversely, the target encoder can use the positional information of the context window to adaptively modulate the type of target features that are feasible to predict for the context encoder module. 
Our proposed conditioning allows the context and target encoders to adapt the set of predictive features based on the size of context or target windows and/or their distance of separation.   
Such ``conditional'' encoders, we term \textbf{E}ncoder \textbf{C}onditioned JEPAs (EC-JEPAs), when used as a drop-in replacement in IJEPA \citep{ijepa} lead to ---
i) improved representational quality measured by rank-based metrics (\emph{LiDAR} \citep{thilak2024lidar} and \emph{RankMe} \citep{garrido_rankme}) as well as classification performance on benchmark datasets such as ImageNet \citep{imagenet} (see \Cref{tab:imagenet1k:compare}), out-of-distribution datasets such as CIFAR10, CIFAR100, Food101 etc. 
ii) improved robustness to context window hyperparameters during pretraining (see \Cref{fig:context-window-abl}) crucial to prevent representational collapse during pretraining 
iii) improved sample-efficiency in pretraining measured by classification performance on ImageNet \citep{imagenet} (see \Cref{fig:pretraining-efficiency}).
\looseness=-1

\section{Method}
\label{sec:main:method}
We first review the IJEPA model \citep{assran2023self} followed by our proposed modification to the same.
\paragraph{IJEPA} 
Let $x \in \R^{T\times d }$ and $p \in \R^{T\times d }$ denote the tokenized input image and position embeddings respectively, where $T$ is the number of tokens, and $d$ the token dimension (we assume position embeddings $p$ are added to the image tokens to produce $x$). 
Let $c$ denote a set of indices corresponding to the context tokens, such that $x_c = \{x_j\}_{j \in c}$. Likewise, let $t^1,..., t^k$ denote $k$ sets of indices with cardinality $m = |t_1| = |t_2|= ... |t_k|$ corresponding to the target token blocks (we use $k=4$ in our experiments following IJEPA \citep{assran2023self}). 
In the IJEPA formulation, an encoder function encodes the context tokens into latent representations $z_c = f(x_c;\theta)$ where $\theta$ are the encoder weights, which are then used to predict the target representations $z_{t^j} = f(x;\tilde{\theta})_{t^j}$ for $j=\{1, ..., k\}$, where $\tilde{\theta}$ are an exponential moving average of the weights $\theta$, with the aid of a predictor function $g$. 
The predictor function takes as input the context representations $z_c$, the target positions $p_{t^j}$, and predicts the targets representations $\hat{z}_{t_j} = g(z_c,p_{t^j};\psi)$ for $j=\{1, ..., k\}$ where $\psi$ are the predictor weights. 

\paragraph{EC-IJEPA} 
In our approach, we use the context and target positions to condition the encoders for pretraining. 
Namely $z_c^{t^1,...,t^k} = f(x_c, p_{t^1},...,p_{t^k};\theta)$, and similarly $z_{t^j}^c = f(x,p_c;\tilde{\theta})_{t^j}$ for $j=\{1,...,k\}$. 
At inference, we simply condition the encoder on all position embeddings $p$. 
In practice, the functions $f$ and $g$ are instantiated as Vision Transformers (ViTs) \citep{dosovitskiy2021vit}, and are conditioned by appending the positions as additional tokens in the input sequence processed by the Transformer modules. 
This increase in sequence length however, could incur a non-negligible cost in memory and compute resources, especially during inference which now processes twice as many tokens as the baseline IJEPA. 
To reduce this computational and memory overhead, we introduce an aggregation step prior to conditioning. 
At both training and inference, we first reduce the conditioning position tokens to a smaller set, which are used as the conditioning tokens instead of the full sequence. 
Concretely, we use 1D average pooling on $p_c, p_{t_1},..., p_{t_k}$ with a kernel and step size of $m//2$\footnote{Note that the target cardinality $m$ is sampled out of a range as in IJEPA}. 
During inference, we use 2D average pooling on all positions $p$ with a kernel and stride size of $[4,4]$. 
This incurs an additional $T//16$ tokens to be processed at inference. 
Finally, we note that we use 1D, rather than 2D average pooling in training due to efficiency and implementation considerations, resulting in approximately $3\%$ increase in FLOPs for training.

\section{Results}
\begin{wraptable}{r}{5.3cm}
\vspace{-13mm}
\caption{\label{tab:imagenet1k:compare} Classification performance comparison on IN-1k dataset.}
\centering
\begin{tabular}{c c}
\toprule
Model & Accuracy \\
\midrule
IJEPA (ViT-L/16)     & 74.8 \\
EC-IJEPA (ViT-L/16)  & \textbf{76.7} \\
\midrule
IJEPA (ViT-H/14)     & 77.4 \\
EC-IJEPA (ViT-H/14)  & \textbf{78.1} \\
\bottomrule
\end{tabular}
\vspace{-4mm}
\end{wraptable}
We evaluate the baseline IJEPA and our proposed encoder conditioned variant EC-IJEPA on several visual benchmarks consistent with prior work \citep{assran2023self, bar2023stochastic}. 
We follow the setup from ~\citet{assran2023self} to pretrain the baseline IJEPA and our proposed EC-IJEPA on the ImageNet-1k (IN-1k) dataset \citep{imagenet} (see \Cref{app:exp-details} for more details). 
The pretrained encoders are then used to extract representations, by average pooling the output sequence of patch-level tokens from the encoder. 
We evaluate these representations on various downstream benchmark datasets using the linear probing protocol adopted by prior work \citep{assran2023self, goyal2021vissl} (see \Cref{app:exp-details} for more details). 


\Cref{tab:imagenet1k:compare} shows the performance of IJEPA and EC-IJEPA on the IN-1k classification benchmark. 
We see that EC-IJEPA outperforms the baseline IJEPA with different encoder sizes.

\begin{wraptable}{r}{7.2cm}
\vspace{-3mm}
\caption{\label{tab:rank_eval} RankMe and LiDAR scores for models pretrained on IN-1k. ViT-L/16 and ViT-H/14 encoders have embedding sizes 1024 and 1280 respectively.}
    \begin{tabular}{c c c}
        \toprule
        Architecture & RankMe $\uparrow$ & LiDAR $\uparrow$ \\
        \midrule
        IJEPA (ViT-L/16) & 488.6 & 385.2 \\
        EC-IJEPA (ViT-L/16) & \textbf{533.0} & \textbf{486.5} \\
        \midrule
        IJEPA (ViT-H/14) & 540.8 & 437.2 \\
        EC-IJEPA (ViT-H/14) & \textbf{567.3} & \textbf{547.0} \\
        \bottomrule
    \end{tabular}
\vspace{-1mm}
\end{wraptable}

Prior works \citep{thilak2024lidar, garrido_rankme} introduced metrics for measuring representational quality that correlate with downstream task performance without the need for a downstream task. 
\emph{RankMe} \citep{garrido_rankme}, is one such metric that measures the soft effective rank of embeddings.
\emph{LiDAR} \citep{thilak2024lidar} is another that builds on \emph{RankMe} by defining a surrogate task to estimate the effective rank of a Linear Discriminant Analysis matrix. 
Both \emph{RankMe} and \emph{LiDAR} metrics empirically show that they serve as useful proxies of representational quality.
Higher scores of these metrics are positively correlated and serve as a necessary condition for improved downstream performance for a given encoder architecture. 
We follow the setup from \citet{garrido_rankme} and \citet{thilak2024lidar} including dataset size and construction to compute these metrics. 
\Cref{tab:rank_eval} shows the \emph{RankMe} and \emph{LiDAR} metrics for IJEPA and EC-IJEPA pretrained on IN-1k. 
We see that EC-IJEPA shows higher scores for \emph{RankMe} and \emph{LiDAR} metrics compared to IJEPA which support the improvements in downstream task performance shown in  \Cref{tab:imagenet1k:compare}.
\looseness=-1

\begin{wrapfigure}{r}{6cm}
    \vspace{-4mm}
    \centering
    \includegraphics[width=\linewidth]{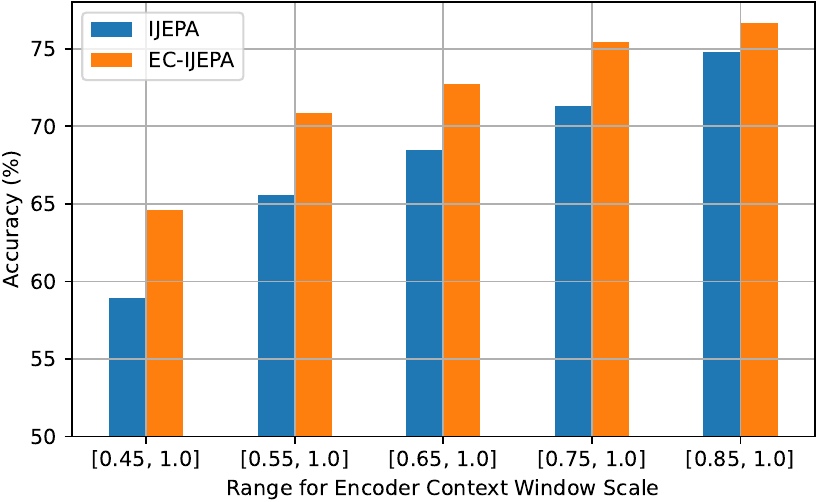}
    \caption{Ablation on ranges of context window scale used for pretraining.}
    \label{fig:context-window-abl}
    \vspace{-5mm}
\end{wrapfigure}

Further, we measure the robustness of the baseline IJEPA and our variant EC-IJEPA to varying sizes for the context window.
\Cref{fig:context-window-abl} compares the classification scores of the baseline and our variant on IN-1k when pretrained for masked prediction task using a wider range of context window sizes using a ViT-L/16 encoder. 
We see that the quality of representations learned by the baseline IJEPA is very sensitive to this hyperparameter. 
In contrast, our variant EC-IJEPA is more robust to a wider range of context window sizes used for masking during pretraining. 
This suggests that our simple positional conditioning alleviates representational collapse in the encoders.

\begin{figure*}[h]
    \vspace{-2mm}
    \centering
    \includegraphics[width=0.75\linewidth]{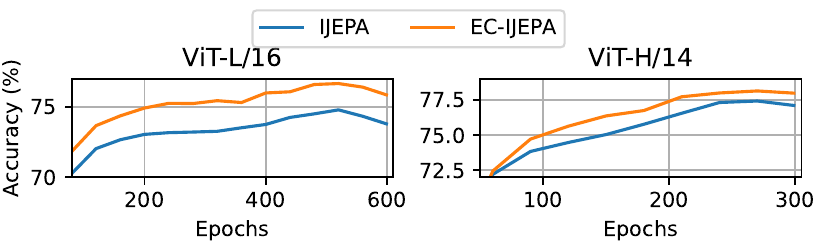}
    \caption{Classification performance on ImageNet-1k measured during pretraining cycle in IJEPA (blue) and EC-IJEPA (orange) at two encoder sizes (left: ViT-L/16 and right: ViT-H/14).}
    \label{fig:pretraining-efficiency}
\end{figure*}

\Cref{fig:pretraining-efficiency} shows the classification accuracy obtained by the baseline IJEPA and our variant EC-IJEPA on IN-1k over the pretraining cycle.
We see that our EC-IJEPA is more sample-efficient for representation learning as it obtains consistently higher classification accuracy throughout the pretraining cycle.
\looseness=-1

\Cref{tab:ood:compare} shows the classification performance of IJEPA and EC-IJEPA on various out-of-distribution datasets such as CIFAR10, CIFAR100, EuroSat, Food101 and SUN397. 
We see that EC-IJEPA consistently outperforms IJEPA which highlights the superior representational quality of the former.
\Cref{tab:ood:compare} also compares performance of models on tasks which require local information such as object counting (CLEVR/Count) and depth prediction (CLEVR/Dist) \citep{johnson2017clevr, zhai2019large} where the two models are comparable with one exception (ViT-L/16 encoder on CLEVR/Dist).
\vspace{-2mm}

\begin{table}[h]
\caption{\label{tab:ood:compare} Classification performance on out-of-distribution datasets using two encoder sizes.}
\centering
\begin{adjustbox}{max width=\textwidth}
\begin{tabular}{c c c c c c c c c}
\toprule
Model & CIFAR10 & CIFAR100 & EuroSat & Food101 & SUN397 & CLEVR/Count & CLEVR/Dist \\
\midrule
IJEPA (ViT-L/16)   & 92.5 & 75.0 & \textbf{96.7} & 75.3 & 69.5 & 74.5 & \textbf{65.3} \\
EC-IJEPA (ViT-L/16) & \textbf{93.4} & \textbf{76.7} & 95.7 & \textbf{76.5} & \textbf{71.2} & \textbf{75.2} & 60.0 \\
\midrule
IJEPA (ViT-H/14) & 94.5 & 78.9 & \textbf{96.5} & 78.4 & 71.5 & \underline{79.3} & \underline{64.8} \\
EC-IJEPA (ViT-H/14) & \textbf{96.0} & \textbf{81.8} & 96.0 & \textbf{78.7} & \textbf{73.5} & \underline{79.4} & \underline{64.6} \\
\bottomrule
\end{tabular}
\end{adjustbox}
\end{table}

\section{Conclusion}
Predictability of patch-level features in natural images has a strong spatial bias. 
We introduce a simple modification to the sequence of input tokens given to the encoder modules in JEPAs, we concatenate positions of target and context windows to the context and target encoders respectively. 
Using our ``conditional'' encoders as a drop-in replacement in IJEPA \citep{assran2023self} shows improved representational quality for downstream image classification tasks and rank-based metrics (\emph{RankMe} and \emph{LiDAR}). 
Conditional encoders alleviate representational collapse across larger ranges of context window sizes and improve sample-efficiency during pretraining.
\looseness=-1

\bibliographystyle{unsrtnat}
\bibliography{reference}

\newpage
\appendix

\section{Experimental Details}
\label{app:exp-details}
\paragraph{Architecture Details.}

We instantiate the context, target and predictor modules in both IJEPA and EC-IJEPA models as Vision Transformers (ViTs)~\citep{dosovitskiy2021vit}. 
We experiment with two different model sizes for the encoder modules, i.e. ViT-Large and ViT-Huge, and a lower capacity ViT Predictor following IJEPA~\citep{assran2023self}.
\Cref{tab:arch:vit_encoder} and \Cref{tab:arch:vit_predictor} respectively show the relevant architecture hyperparameters for the ViT-based encoders and predictors.

\begin{table}[htbp]
    \centering
    \caption{Encoder architecture using ViT-based models. The value after ``/'' indicates the patch size.}
    \begin{tabular}{c c c c}
        \toprule
         Architecture & Depth & Hidden Dimension & Number of Heads\\
         \midrule
         ViT-L/16 & 24 & 1024 & 16 \\
         ViT-H/14 & 32 & 1280 & 16 \\
         \bottomrule
    \end{tabular}
    \label{tab:arch:vit_encoder}
\end{table}

\begin{table}[htbp]
    \centering
    \caption{Predictor architecture using ViT-based models. Number of heads is set to match that of the encoder.}
    \begin{tabular}{c c c c}
        \toprule
         Architecture & Depth & Hidden Dimension & Number of Heads \\
         \midrule
         ViT-Predictor & 12 & 384 & 16 \\
         \bottomrule
    \end{tabular}
    \label{tab:arch:vit_predictor}
\end{table}

\paragraph{Pretraining Details.}
We use the AdamW optimizer \citep{loshchilov2018decoupled} ~\footnote{https://pytorch.org/docs/stable/generated/torch.optim.AdamW.html} to train IJEPA and EC-IJEPA in all our experiments. ~\Cref{tab:optim:vit_large} and~\Cref{tab:optim:vit_huge} show the hyperparameters used to pretrain all models in this work. 
We follow the pretraining configuration from IJEPA ~\citep{assran2023self}. 
We follow masking hyperparameters used to create context and target masks from IJEPA \citep{assran2023self}.

\begin{table}[htbp]
    \centering
    \caption{Pretraining hyperparameters used for ViT-L/16}
    \label{tab:optim:vit_large}
    \begin{tabular}{c|c}
      \toprule
      Hyperparameter & Value \\
      \midrule
      Optimizer  &  AdamW \\
      Epochs  &  600 \\
      Max learning rate  &  0.001 \\
      LR Warmup type & Linear \\
      LR Decay type & Cosine \\
      Warmup epochs & 15 \\      
      Batch size & 2048 \\
      Weight decay scheduler  &  Cosine \\
      Weight decay (start, end) &  [0.04, 0.4] \\
      EMA momentum scheduler & Linear \\
      EMA momentum (start, end)  &  [0.996 1.0] \\
      \bottomrule
    \end{tabular}
\end{table}

\begin{table}[htbp]
    \centering
    \caption{Hyperparameter configuration used to pretrain ViT-H/14}
    \label{tab:optim:vit_huge}
    \begin{tabular}{c|c}
      \toprule
      Hyperparameter & Value \\
      \midrule
      Optimizer  &  AdamW \\
      Epochs  &  300 \\
      Max learning rate  &  0.001 \\
      LR Warmup type & Linear \\
      LR Decay type & Cosine \\
      Warmup epochs & 40 \\
      Batch size & 2048 \\
      Weight decay scheduler & Cosine \\
      Weight decay (start, end) & [0.04, 0.4] \\
      EMA momentum scheduler & Linear \\
      EMA momentum (start, end)  &  [0.996 1.0] \\
      \bottomrule
    \end{tabular}
\end{table}

\paragraph{Evaluation on ImageNet-1k}

We evaluate the pretrained encoders described above using linear probing on ImageNet-1k dataset~\citep{imagenet}. We adapt the evaluation protocol from IJEPA~\citep{assran2023self} wherein the pretrained model weights are frozen and are used to extract a feature vector by average pooling (across the sequence length) the output tokens from the last layer of the encoder. 
A linear probe that consists of a batch normalization layer with non-learnable affine parameters followed by a linear layer is used to map this feature vector to the set of classification logits on ImageNet-1k dataset. 
The parameters of the linear probe are trained with the LARS~\citep{LARSYou2017} optimizer using a learning rate of $0.05$, no weight decay and with a batch size of $16384$ for $50$ epochs.

\paragraph{Evaluation on out-of-distribution (OOD) datasets}

We use CIFAR10, CIFAR100~\citep{cifar}, EuroSAT~\citep{eurosat}, Food101~\citep{food101}, SUN397~\citep{sun397}, CLEVR/Count and CLEVR/Dist~\citep{johnson2017clevr, zhai2019large} as unseen or OOD datasets w.r.t the pretraining dataset (ImageNet-1k). 
We again adopt the evaluation protocol of linear probing with a frozen backbone. 
We follow the evaluation protocol used in VISSL~\citep{goyal2021vissl} also used in prior works \citep{assran2023self, bar2023stochastic} to train and evaluate a linear probe for the OOD datasets.
~\Cref{tab:ood:optim} lists the relevant hyperparameter configurations used in our experiments.


\begin{table}[h]
    \centering
    \caption{Hyperparameters used for linear evaluation on OOD datasets.}
    \label{tab:ood:optim}
    \begin{tabular}{c c c c c c c}    
        \toprule
        Dataset & Optimizer & Momentum & Weight & Learning rate & Epochs \\
                &           &          &  decay &  (LR)         &        \\
        \midrule
        CIFAR10     & SGD with Nesterov & 0.9 & 0.0005 & 0.01 & 28  \\
        CIFAR100    & SGD with Nesterov & 0.9 & 0.0005 & 0.01 & 28  \\
        EuroSAT     & SGD with Nesterov & 0.9 & 0.0005 & 0.01 & 28  \\
        Food101     & SGD with Nesterov & 0.9 & 0.0005 & 0.01 & 28  \\
        SUN397      & SGD with Nesterov & 0.9 & 0.0005 & 0.01 & 28  \\
        CLEVR/Count & SGD with Nesterov & 0.9 & 0.0005 & 0.01 & 50  \\
        CLEVR/Dist  & SGD with Nesterov & 0.9 & 0.0005 & 0.01 & 50  \\
        \bottomrule
    \end{tabular} 
\end{table}

\section{Additional Results}
\begin{figure}
    \centering
    \includegraphics[width=0.75\textwidth]{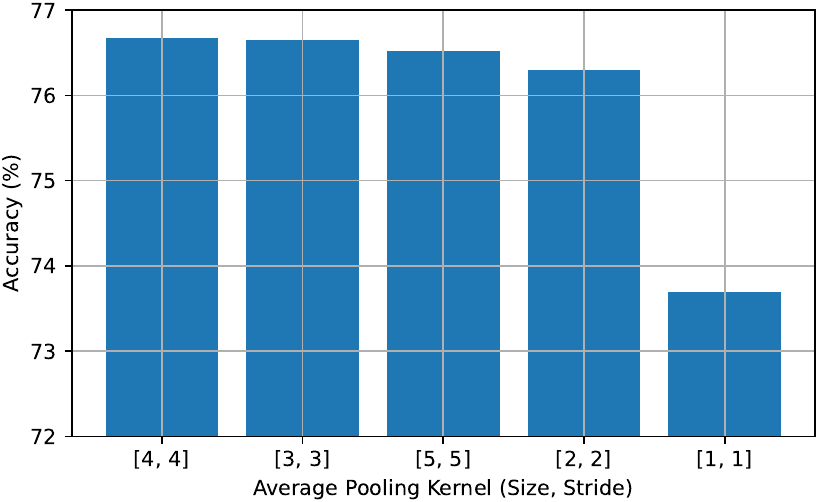}
    \caption{Linear probing accuracy on Imagenet-1k dataset w.r.t kernel size and stride.}
    \label{fig:app:inference_kernel_ablation}
\end{figure}

\paragraph{Average Pooling Ablation.}
EC-IJEPA uses average pooling with a kernel size and stride of $[4, 4]$ respectively at inference time to create conditioning position tokens as described in ~\Cref{sec:main:method}. 
We perform an ablation experiment to measure the impact of kernel size and stride on downstream classification accuracy on ImageNet-1k~\citep{imagenet} by varying these hyperparameters.
~\Cref{fig:app:inference_kernel_ablation} shows the maximum classification accuracy achieved on ImageNet-1k validation as a function of kernel size and stride. 
We observe from~\Cref{fig:app:inference_kernel_ablation} that the highest accuracy is achieved with a kernel size of $4$ and stride of $4$. 
Furthermore, we observe that there is a drop off in accuracy for kernel size of $1$ and stride of $1$. 
These observations suggest that the values for these hyperparameters used in~\Cref{sec:main:method} are reasonable to extract representations from EC-IJEPA for classification tasks.

\end{document}